\documentclass[10pt,twocolumn]{article} 
\usepackage{simpleConference}
\usepackage{times}
\usepackage{graphicx}
\usepackage{amssymb}
\usepackage{url,hyperref}
\usepackage{cite}
\usepackage{amsmath,amssymb,amsfonts}
\usepackage{algorithmic}
\usepackage{graphicx}
\usepackage{textcomp}
\usepackage{xcolor}
\usepackage[numbers]{natbib}
\usepackage{tabularx}
\usepackage{graphicx}
\usepackage{multirow}
\usepackage{subcaption}
\usepackage{amsmath} 

\begin{document}

\title{A Safe Exploration Strategy for Model-free Task Adaptation  in Safety-constrained Grid Environments}

\author{
    \small
    \begin{tabular}{c}
        \textbf{Erfan Entezami} \\
        \textit{Department of Computer Sciences} \\
        \textit{University of Massachusetts Amherst} \\
        Amherst, MA, USA \\
        \texttt{eentezami@cs.umass.edu}
    \end{tabular}
    \hspace{0.5em} 
    \begin{tabular}{c}
        \textbf{Mahsa Sahebdel} \\
        \textit{Department of Computer Sciences} \\
        \textit{University of Massachusetts Amherst} \\
        Amherst, MA, USA \\
        \texttt{msahebdelala@umass.edu}
    \end{tabular}
    \hspace{0.5em} 
    \begin{tabular}{c}
        \textbf{Dhawal Gupta} \\
        \textit{Department of Computer Sciences} \\
        \textit{University of Massachusetts Amherst} \\
        Amherst, MA, USA \\
        \texttt{dgupta@cs.umass.edu}
    \end{tabular}
}
\maketitle
\thispagestyle{empty}

\begin{abstract}
Training a model-free reinforcement learning agent requires allowing the agent to sufficiently explore the environment to search for an optimal policy. In safety-constrained environments, utilizing unsupervised exploration or a non-optimal policy may lead the agent to undesirable states, resulting in outcomes that are potentially costly or hazardous for both the agent and the environment. In this paper, we introduce a new exploration framework for navigating the grid environments that enables model-free agents to interact with the environment while adhering to safety constraints. Our framework includes a pre-training phase, during which the agent learns to identify potentially unsafe states based on both observable features and specified safety constraints in the environment. Subsequently, a binary classification model is trained to predict those unsafe states in new environments that exhibit similar dynamics. This trained classifier empowers model-free agents to determine situations in which employing random exploration or a suboptimal policy may pose safety risks, in which case our framework prompts the agent to follow a predefined safe policy to mitigate the potential for hazardous consequences. We evaluated our framework on three randomly generated grid environments and demonstrated how model-free agents can safely adapt to new tasks and learn optimal policies for new environments. Our results indicate that by defining an appropriate safe policy and utilizing a well-trained model to detect unsafe states, our framework enables a model-free agent to adapt to new tasks and environments with significantly fewer safety violations.
\end{abstract}

\begin{figure*}[htbp]
    \centering
    \includegraphics[width=0.8\textwidth]{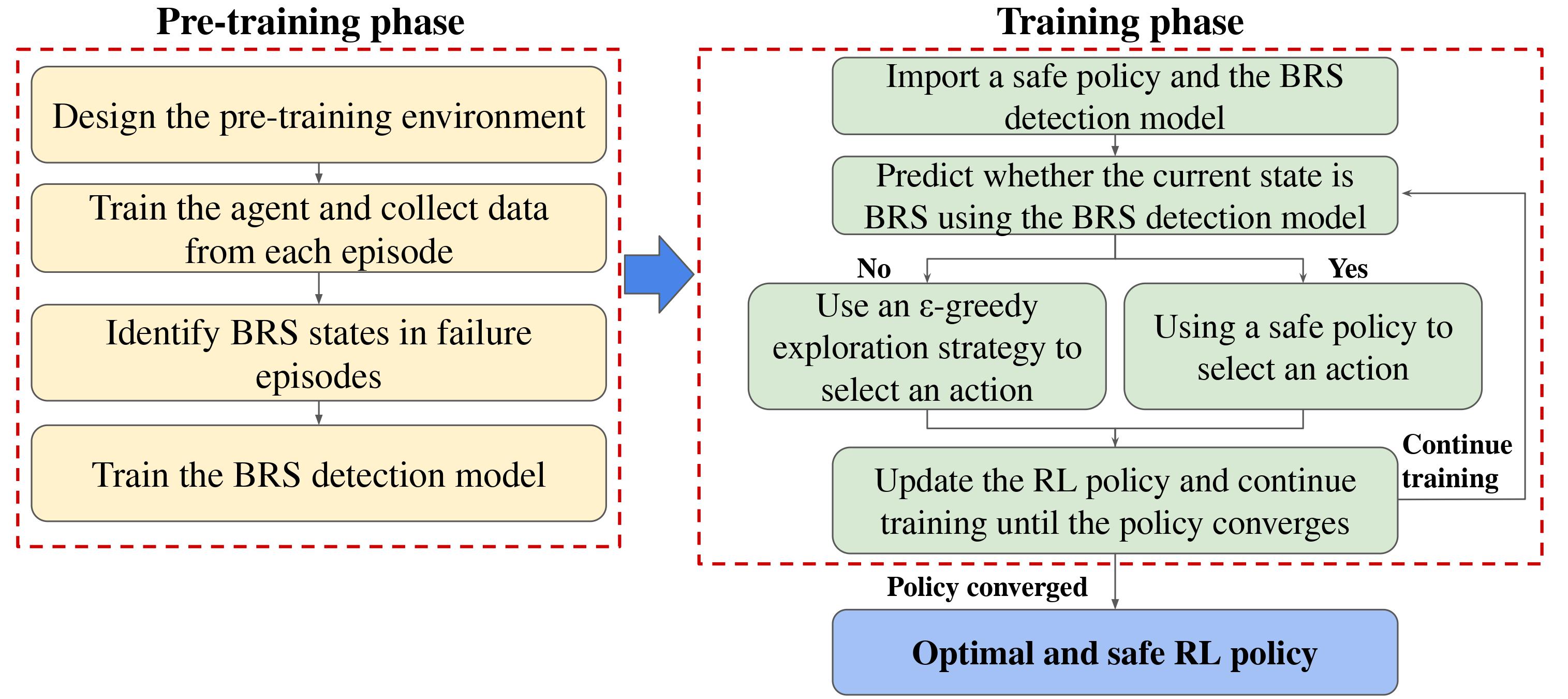}
    \caption{During the pre-training phase, a binary classification model is trained using features extracted from both BRS and non-BRS states. This model is subsequently used in a new environment to identify situations where using $\epsilon$-greedy exploration strategy might pose a risk.}
    \label{fig:fig1}
\end{figure*}

\section{Introduction}
Learning through trial and error is a critical approach for humans to learn new concepts which has also been an inspiration for developing \textit{Reinforcement Learning} (RL) methods~\citep{sutton2018reinforcement}. In RL methods, the desired behavior is learned through interaction with the environment, where the agent receives feedback based on the decisions it makes. While RL methods have demonstrated remarkable performance across various domains, including video games~\citep{peng2017multiagent, lee2018modular, torrado2018deep}, conventional recommender systems~\citep{zou2019reinforcement, tang2019reinforcement, ji2021reinforcement}, and enhancing the responses of large language models~\citep{dai2023safe, ouyang2022training}. Their applications in many real-world situations such as autonomous driving~\citep{kiran2021deep,baheri2020deep}, medical usage~\citep{sui2017correlational,sui2015safe, thomas2019preventing} and recommendation systems in sensitive domains~\citep{sui2015safe, singh2020building} have been restricted due the potential to make dangerous and irreversible mistakes. This problem is more severe for model-free agents given that they don't have any prior knowledge about the environment and they predominantly rely on a trial-and-error approach to learn the optimal policy.

In numerous real-world scenarios, the environment within which the agents interact demands careful consideration of safety. For instance, in the realm of autonomous cars, many safety features have been developed to identify possible collision situations and take preventive actions to avoid them. In the field of medical applications, providing suggestions for intricate and potentially dangerous situations often involves discussions led by professional experts, rather than relying solely on AI-based systems. Similarly, recommendation systems on movie platforms must consciously account for safety considerations, ensuring the appropriateness of suggestions for specific age ranges. In all the examples provided, diverse solutions for safety considerations have been implemented to identify hazardous states and prompt alternative responses beyond their initially planned actions. These responses may involve reducing the speed of the vehicle to avoid a collision, deferring diagnosis to experts instead of relying solely on AI, or executing predefined tasks such as excluding certain suggestions from the available options in a recommendation system.

Applying RL in real-world scenarios that demand safety considerations, similar to the examples mentioned earlier, introduces the risk of potentially dangerous mistakes due to the random exploration required during the training phase ~\citep{sui2015safe, garcia2012safe, hans2008safe}. On the other hand, exploration is an indispensable component of RL, allowing the agent to learn the optimal policy, especially in model-free settings where no information about the task or environment is provided to the agent. The inherent risk associated with the use of RL in these scenarios, arising from the free exploration strategy, makes employing RL an inefficient approach.

In this work, we propose a framework for training model-free RL agents to navigate grid environments where exploration requires consideration of safety constraints. These environments encompass hazardous states, the exploration of which might be dangerous or costly for the agent or the environment. Consequently, employing RL in such contexts amplifies the potential of encountering undesirable states due to unsupervised exploration strategies.

Our method empowers the RL agent to identify risky states, where the agent may navigate towards undesirable states through inappropriate actions selected by random exploration or a suboptimal policy. Subsequently, our framework enables the agent to recognize situations where free exploration becomes unsafe, prompting the selection of actions based on a safe policy rather than relying on random or suboptimal policies. Defining the safe policy is highly dependent on the environment and can be a set of predefined actions for the agent, relinquishing control to a human or supervisor, or employing any other secure approach as a substitute for the suboptimal RL policy that utilizes a free exploration strategy. The primary contribution of this paper lies in the design of a framework that can efficiently detect situations when free interaction (random exploration or using suboptimal policy) is unsafe for the model-free RL agent. The proposed method determines when free interaction should be replaced with a more reliable and secure strategy for selecting actions to learn the optimal policy while reducing the risk of visiting undesirable states in the training phase.

As shown in Figure~\ref{fig:fig1} our proposed framework involves a pre-training phase conducted in a simulator or controlled segment of the environment, that has the same dynamics as the main environment but without hazardous consequences for entering undesirable states. During the pre-training phase, the agent interacts freely with the pre-training environment to train a model capable of identifying states that may not be initially undesirable but if the agent enters these states, inappropriate actions selected by random exploration or suboptimal policy may lead the agent into undesirable states. The model trained in the pre-training phase is then applied to the main environment, which the model-free agent has not encountered before. This enables the agent to recognize unsafe situations, where free interaction with the environment is risky. In such cases, future actions are selected using a safe policy which is defined particularly for each environment and task.

\section{Related Work}

Given the inherent risk and uncertainty associated with applying RL methods, safety considerations have been extensively investigated within the realm of RL methods \citep{brunke2022safe, garcia2015comprehensive, garcia2012safe, fulton2018safe, thomas2019preventing, turchetta2016safe, zhang2020cautious}.

One of the most important safety considerations in RL revolves around safe exploration ~\citep{sui2015safe, garcia2012safe,mannucci2017safe, gillulay2011guaranteed, geibel2001reinforcement, pecka2014safe, liu2023safe} to avoid exposing the agent to dangerous states and consequently avoid harmful outcomes.
Abeel et al. \cite{abbeel2005exploration} proposed an apprenticeship framework to train an RL agent in safety-critical environments where the agent relies on a teacher that should act safely and near-optimally. Hans et al.~\cite{hans2008safe} introduced a method to assess a state's safety level, alongside a backup policy to navigate the system from critical to safe states. In \cite{moldovan2012safe} 
 the need for a safe exploration strategy in \textit{Markov Decision Process} (MDP) has been discussed, and a safe but potentially suboptimal exploration strategy for safety-constrained environments has been proposed. Turchetta et al. \cite{turchetta2016safe} proposed SAFEMDP algorithm which is more similar to our work and enables RL agents to explore the reachable portion of the MDP while adhering to safety constraints and gaining statistical confidence in unvisited state-action pairs using noisy observations.

Unlike many similar works on safe exploration techniques, our framework does not require prior knowledge about the distribution of unsafe states or initializing in a safe state. However, our framework needs pre-training in an environment with similar dynamics to the main environment in order to learn risky situations, given information provided in the state representation and/or agent observations.

Safe training for RL methods in safety-constrained environments has been extensively explored \cite{zhang2023evaluating,zhao2021model, thananjeyan2021recovery, zhao2023state, alshiekh2018safe}. As a few examples, in \cite{zhao2021model} a model-free safe control strategy for deep reinforcement learning (DRL) agents has been suggested to prevent safety violation in the training phase. Thananjeyan et al. \cite{thananjeyan2021recovery} proposed an algorithm called Recovery RL that leverages offline data in order to identify safety violation zones before policy learning and similar to our method, utilizes two different policies, one for learning the task and another for considering the safety constraints in the environment. Our framework proposes a more structured strategy to identify dangerous situations by computing reachable sets of undesirable states using high-level features that are observable for the agent. Another difference between our framework and methods similar to Recovery RL lies in defining the backup (safe) policy. Unlike these methods, our approach leverages a predefined safe policy instead of learning it from offline data. We propose defining a predefined safe policy and primarily focus on identifying dangerous situations in previously unseen environments with different state spaces and reward distributions, rather than learning the safe policy and applying it to a similar environment for a different task.

Another set of methods similar to ours addresses safety in reachability problems, aiming to enable RL agents to learn safe policies considering environmental safety constraints. Similar approaches \citep{fisac2019bridging, fisac2018general, hsu2021safety, li2022infinite} have proposed enabling RL agents to attain a safe policy by computing reachable sets using Hamiltonian methods and defining a value function to assign negative values to reachable areas of undesirable states. Some key differences between these methods and our framework lie in applications, computational complexity, and the primary goal of the methods. Our method focuses on providing a safe learning framework, whereas other methods primarily analyze the safety levels of learned policies.

\begin{figure}[htbp]
    \centering
    \includegraphics[width=\columnwidth]{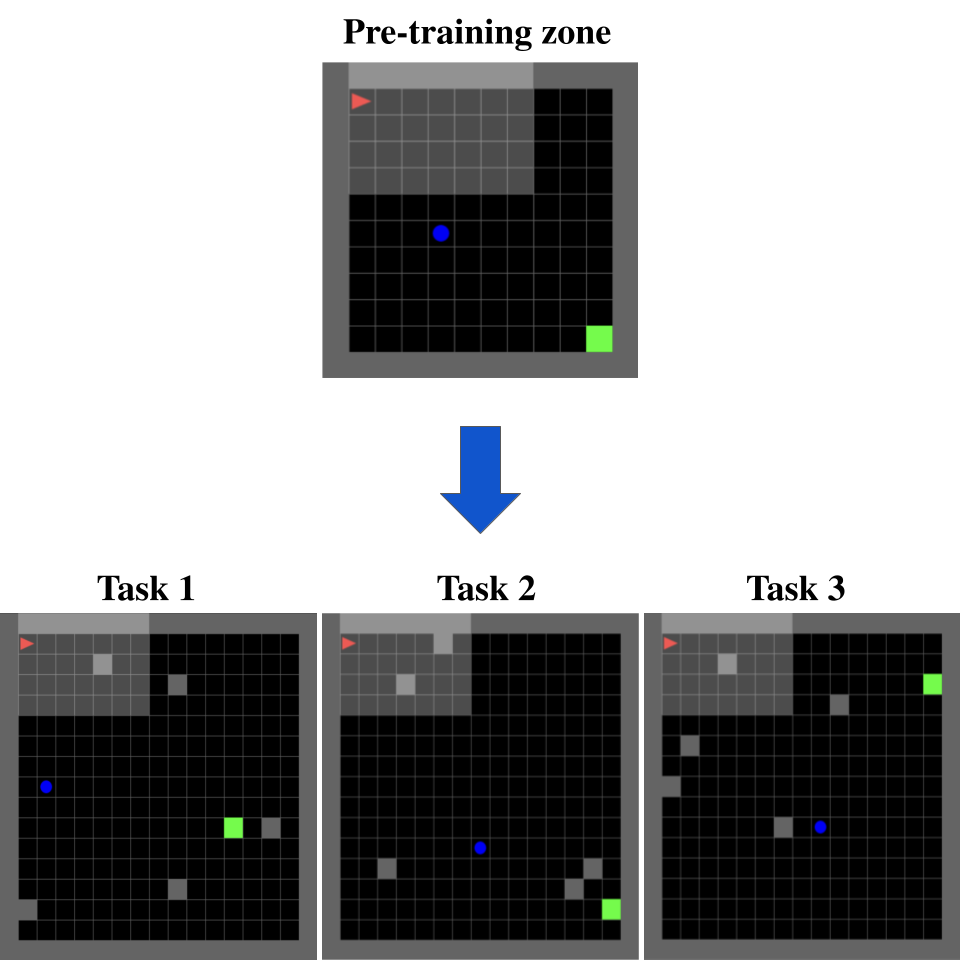}
    \caption{We designed a 10x10 grid environment (Pre-training zone) containing one moving obstacle and one goal state to train the BRS detection model, and we created three 15x15 randomly generated grid environments (Task 1 to 3), each containing one moving obstacle, five blocked states, and one goal state to evaluate the performance of our framework.}
     \label{fig:main}
\end{figure}

\section{Problem Definition and Background}
Similar to most RL problems, we formulate our problem of interest as a \textit{Markov Decision Process} (MDP) which is defined as a tuple $(S, A, P, R, \gamma)$, where $S$ is the set of states; $A$ denotes the set of actions; $P(s_{t+1} = s' | s_t=s, a_t=a)$ is the transition function that represents the dynamics of the environment and returns the probability of transitioning to a specific state $s_{t+1} \in S$ given the agent's previous state $s_t \in S$ and the action $a_{t} \in A$ chosen at state $s_t$ state;  $R: S \times A \rightarrow \Re$ is the reward function and $\gamma \in [0, 1]$ is the discount factor.

We denote the set of failure states as $S_{f} \subset S$ which getting into them is dangerous or costly for the agent and/or the environment. The complementary set of $S_{f}$ is $S_{s} \subset S$ which contains all safe states that the agent can interact with to learn the optimal policy for the given task. Our objective is to equip the agent with the capability to identify states that upon transitioning to them, random exploration or utilizing a suboptimal policy may direct the agent into undesirable states $S_{f}$ within the next $t$ timesteps.

To determine these states, we utilize the computation of a set called \emph{Backward Reachable Set} (BRS). BRS is widely used in various domains such as control and reachability problems ~\citep{fisac2018general, bansal2017hamilton}, robot navigation~\citep{bajcsy2019efficient}, and reinforcement learning ~\citep{hsu2021safety,fisac2019bridging}. The BRS is the set of states $S_{BRS} \subset S$ in which the agent can end up within a set of states called target states $S_{target} \subset S$ within the next $t$ timesteps ~\citep{bansal2017hamilton}. In this study, we compute the BRS for our failure states. Thus our set of target states is $S_f$ and in the following sections, $S_{BRS}$ denotes the backward reachable set of failure states $S_f$.

We consider a trajectory of states $\xi_{s_0}^{\pi} = \{s_0, s_1, s_2, ...\}$, to be safe, when it starts from $s_0$ and follows a policy $\pi$ for a maximum of $T$ timesteps if and only if $\forall s_j \in \xi_{s_0}^{\pi} ; s_j \in S_s$ for $j \in [0,T]$. $S_{BRS}$ comprises of all states $s_i \in \xi_{s_0}^\pi$  leading to failure states within $t$ timesteps, i.e., $\exists k \in [0,t],$ s.t. $s_{i+k} \in S_f$. $S_{BRS}$ with respect to $S_f$ over a time horizon of maximum $t$ timesteps is defined as follows:


\begin{align}
\label{eq:equation1}
 S_{BRS} &= \left\{ s_i \in S \, \middle| \, \exists k\leq t,  s_{i+k} \in \xi_{s_0}^{\pi}, \right. \notag \\
 &\qquad \left. s_{i+k}  \in S_f \right\} \quad t\in[0, T]
\end{align}

\section{Method}
Our proposed framework includes two phases, namely pre-training and safe training. In the pre-training phase, a binary classification model is trained to detect dangerous states. In the safe training phase, this model is applied to identify dangerous states in a new environment. Upon detection, the agent switches from its current action selection strategy, which may involve random exploration or a suboptimal policy, to a more reliable policy known as the safe policy to prevent the agent from entering undesirable states.

\subsection{Pre-training Phase}
The pre-training phase takes place in an environment that has similar dynamics as the main environment. It could be either a controlled portion of the main environment or a simulated environment where getting into failure states does not have catastrophic consequences.

The primary goal during the pre-training phase is to train a model capable of predicting the states that lie in BRS given observable features for the agent in the environment. In our specific domain, the safety constraint entails avoiding collisions with a moving obstacle.

We allow the agent to interact with the pre-training environment using an $\epsilon$-greedy exploration strategy to find the optimal policy for a given task, while simultaneously computing the BRS from the episodes where the agent encountered failure states $S_f$. To that end, we define a signed function $l: S \to \Re, \quad l(s) > 0 \iff s \in S_s$ which is a signed distance between the current state and the target state which in our work is an undesirable state. 
Since $l$ is a signed distance function, by using the function $l$, it becomes possible to identify whether the agent reaches failure states $S_f$. 
Using the defined distance function $l(s)$, we consider a conservative perspective to formulate a value function for states in the trajectory $\xi_{s_0}^\pi$ as follows: 

\begin{align}
\label{eq:equation2}
    V(s) = \min_{i \in [0, t], s_i \in \xi_{s_0}^\pi} l(s_i), \quad  s \in \xi_{s_0}^\pi \quad 
\end{align}

The proposed value function is designed to capture the minimum distance with the obstacle that is achieved by a trajectory starting from state $s_0$ by following policy $\pi$ over a time horizon of [0, t]. In other words, this value function represents the minimum value attained by all states in a trajectory starting from $s_0$ and following policy $\pi$ over a time horizon with the duration of $t$ timesteps. This minimum value is then assigned as the value for all states in the trajectory $\xi_{s_0}^{\pi}$. Given the defined value function, if one state in the trajectory is an undesirable state ($s_i \in \xi_{s_0}^\pi \text{ and } s_i \in S_f$), a negative value is assigned not only to that state but also to all other states present in the same trajectory. By computing the proposed value function within a finite time horizon [0, t], we are now able to redefine the BRS as follows:

\begin{align}
\label{ref: final_BRS_value}
S_{BRS} = \{s \in S : V(s) \leq  0\} \quad 
\end{align}

By utilizing the proposed value function, we can identify BRS states and utilize them to train a binary classification model. This model is then employed in new environments to detect whether the agent is in a BRS state.

\subsection{Safe Training Phase}


When the agent encounters a new environment wherein it has to learn a policy to solve the given task, it often searches for optimal policy by utilizing some form of exploration that can
potentially be random, which may lead to a suboptimal policy during the transient search phase for the optimal policy.
By leveraging the BRS detection model, we can enable the agent to identify states within the BRS of undesirable states. This model determines whether the agent is currently within the BRS of undesirable states given the agent's observations. Upon detection of such states, the agent can replace an exploratory or suboptimal policy, which it had been using to explore the environment, with a predefined safe policy. Switching between a suboptimal/random policy and a safe policy when the agent enters $S_{BRS}$ allows the agent to explore the environment while avoiding entering undesirable states $S_f$. The definition of the safe policy is highly dependent on the environment and the task and will be further explained in Section \ref{sec: Discussion}.

\section{Experimental Results}
We evaluated our framework through an autonomous navigation task designed with MiniGrid~\citep{chevalier2023minigrid} platform. As shown in Figure~\ref{fig:main} our designed environments contain a moving obstacle that the agent must navigate around and reach the goal state. The obstacles move vertically in all environments and change direction upon reaching the upper or lower boundary of the environments. Reaching the goal state or interacting with the environment for a duration exceeding a maximum threshold of timesteps terminates the episode with a reward of +1 and 0, respectively. Collision with the moving obstacles constitutes a safety constraint violation in this setting which terminates the episode with a reward of -1. The environment encompasses four actions: "turn right", "turn left", "move forward", and "do nothing", that turning actions alter the agent's direction, while the "move forward" action moves the agent by one state in its current direction. The "do nothing" action is exclusively designated for our safe policy, as outlined later, and is only accessible when the agent employs the safe policy. The objective of each task in this setup is for the agent to learn a policy to reach the goal state without colliding with the moving obstacle. The state representation in this scenario includes the agent's vertical and horizontal positions, as well as the vertical position of the obstacle, (given its exclusive vertical movement) and its direction. QLearning and SARSA, two instances of model-free RL algorithms with distinct characteristics, were examined to assess our framework. However, given that our framework primarily concerns exploration strategy rather than specific learning algorithm properties, any alternative model-free algorithm can be utilized in place of them.

\begin{table*}[ht]
    \centering
    \caption{Comparison of learning the optimal policy using $\epsilon$-greedy and our exploration strategy for QLearning and SARSA algorithms.}
    \label{comparison}
    \begin{tabularx}{\textwidth}{|X|X|p{1.5cm}|p{1.5cm}|p{1.5cm}|p{1.5cm}|p{1.5cm}|p{1.5cm}|}
        \hline
        & & \multicolumn{3}{c|}{$\epsilon$-greedy Exploration} & \multicolumn{3}{c|}{Safe Exploration} \\
        \cline{3-8}
        Algorithm & Task & Average Collision Rate & Average Success Rate & Sum of Reward & Average Collision Rate & Average Success Rate & Sum of Reward \\
        \hline
        \multirow{3}{*}{QLearning} & Task 1 & 0.287 & 0.679 & 405.259 & 0.062 & 0.860 & 1090.833 \\
        & Task 2 & 0.148 & 0.849 & 1082.735 & 0.032 & 0.965 & 1486.630 \\
        & Task 3 & 0.129 & 0.870 & 1313.034 & 0.048 & 0.950 & 1600.665 \\
        \hline
        \multirow{3}{*}{SARSA} & Task 1 & 0.170 & 0.772 & 732.294 & 0.011 & 0.892 & 1217.235 \\
        & Task 2 & 0.101 & 0.895 & 1232.721 & 0.009 & 0.987 & 1559.047 \\
        & Task 3 & 0.092 & 0.906 & 1425.101 & 0.027 & 0.972 & 1670.998 \\
        \hline
    \end{tabularx}
    \label{tab1}
\end{table*}

 \subsection{Pre-training}
We initially define a simpler environment in a 10x10 grid that contains a moving obstacle and a goal state. We train the agent for 4000 episodes in the pre-training environment with a high exploration rate (0.6) and collect data from all training episodes. To identify the BRS for the moving obstacle in the pre-training environment, the signed distance function ($l(s)$) is defined as the Euclidean distance between the obstacle and the agent. We compute the BRS of the moving obstacle for $ t=2$ timesteps utilizing the value function described in equation \ref{ref: final_BRS_value}. By computing BRS during the pre-training phase, we identified states where if the agent enters them, random exploration or actions chosen by a suboptimal policy can potentially lead the agent to collide with the obstacle within the next 2 timesteps. After identifying the BRS states, we train a binary classification model to detect BRS and non-BRS states given state features including the direction of the agent, the direction of the moving obstacle, and the signed distance.

In this work, we proceeded to train several classification models, including KNN, SVM, random forest, decision tree, and a simple CNN, to perform a binary classification task for detecting BRS and non-BRS states. Given the properties of the features present in the state representation and agent observation, the SVM classification model exhibited superior performance compared to other tested methods (when we utilized Q-Learning as our pre-training algorithm, SVM achieved an accuracy of 0.92 and an F1 score of 0.83, whereas it attained an accuracy of 0.92 and an F1 score of 0.77 when the SARSA algorithm was employed). Therefore, the SVM classification model has been chosen as the primary BRS detection model in our experiments.

\begin{figure}[ht]
    \centering
    \begin{subfigure}[b]{0.235\textwidth}
        \centering
        \includegraphics[width=\textwidth]{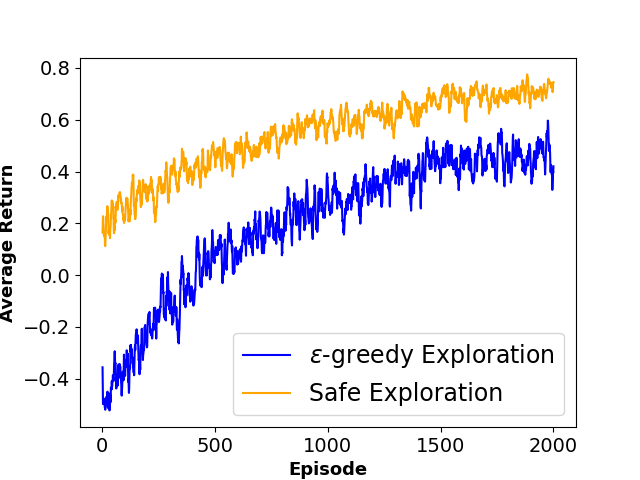}
        \caption{Task 1}
        \label{fig:sub1}
    \end{subfigure}
    \begin{subfigure}[b]{0.235\textwidth}
        \centering
        \includegraphics[width=\textwidth]{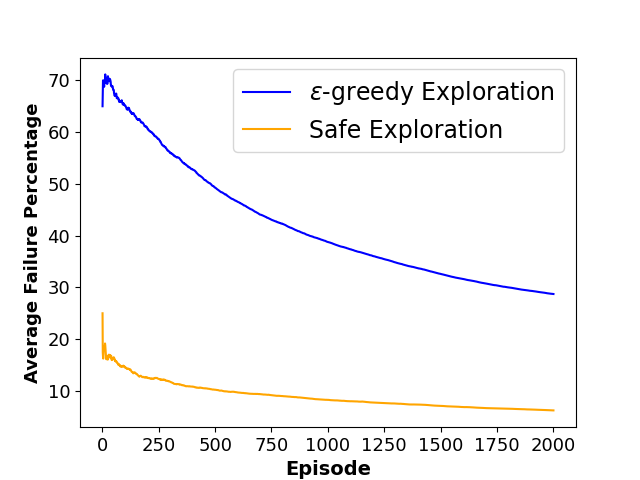}
        \caption{Task 1}
        \label{fig:sub2}
    \end{subfigure}
    \\
    \begin{subfigure}[b]{0.235\textwidth}
        \centering
        \includegraphics[width=\textwidth]{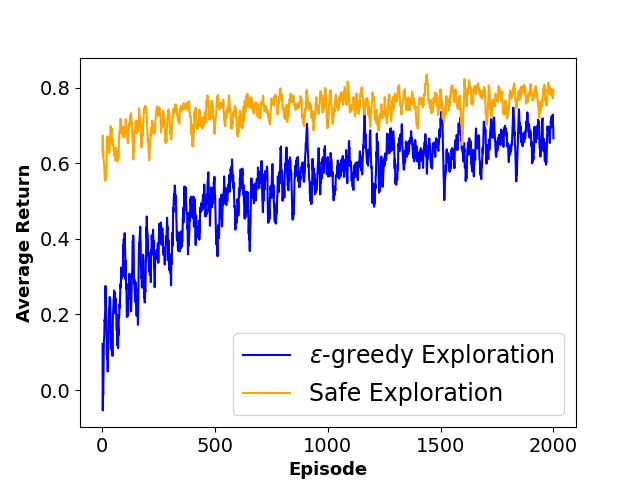}
        \caption{Task 2}
        \label{fig:sub3}
    \end{subfigure}
    \begin{subfigure}[b]{0.235\textwidth}
        \centering
        \includegraphics[width=\textwidth]{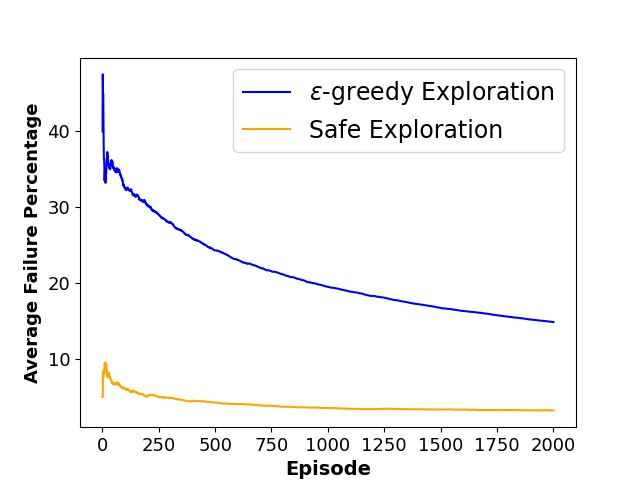}
        \caption{Task 2}
        \label{fig:sub4}
    \end{subfigure}
    \\
    \begin{subfigure}[b]{0.235\textwidth}
        \centering
        \includegraphics[width=\textwidth]{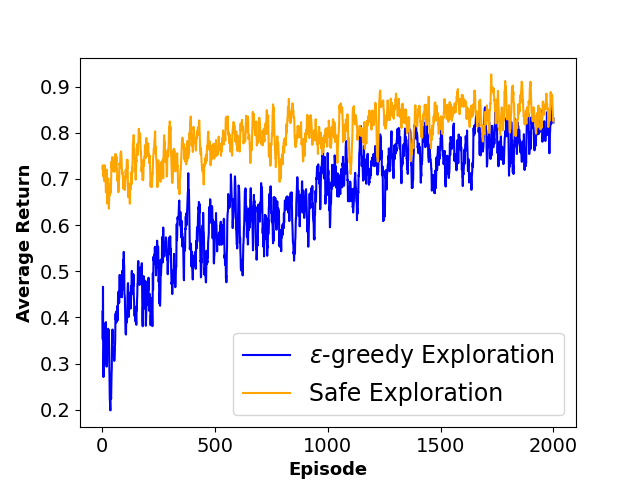}
        \caption{Task 3}
        \label{fig:sub5}
    \end{subfigure}
    \begin{subfigure}[b]{0.235\textwidth}
        \centering
        \includegraphics[width=\textwidth]{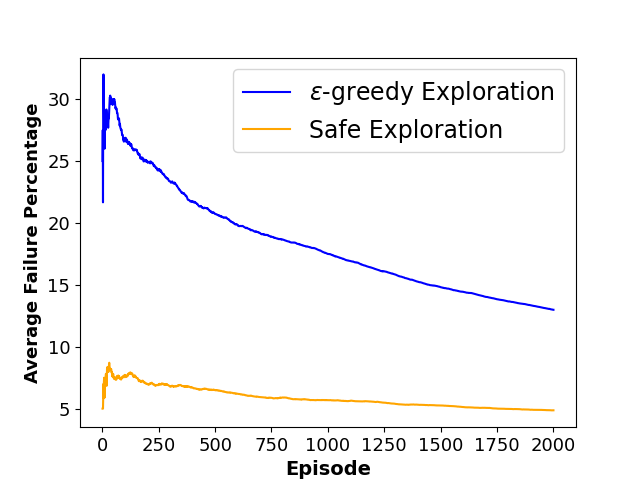}
        \caption{Task 3}
        \label{fig:sub6}
    \end{subfigure}
    \caption{Training process of QLearning algorithm for the designed tasks. Diagrams on the left depict the mean of the average returns for the 10 most recent episodes, and those on the right illustrate the percentage of episodes that ended with collision. All values are obtained by running each test 20 times and getting the average results. In all experiments, we used $\gamma$ = 0.99, exploration rate = 0.2 and learning rate = 0.5 as our hyperparameters.}
    \label{fig:overal1}
\end{figure}

\begin{figure}[ht]
    \centering
    \begin{subfigure}[b]{0.235\textwidth}
        \centering
        \includegraphics[width=\textwidth]{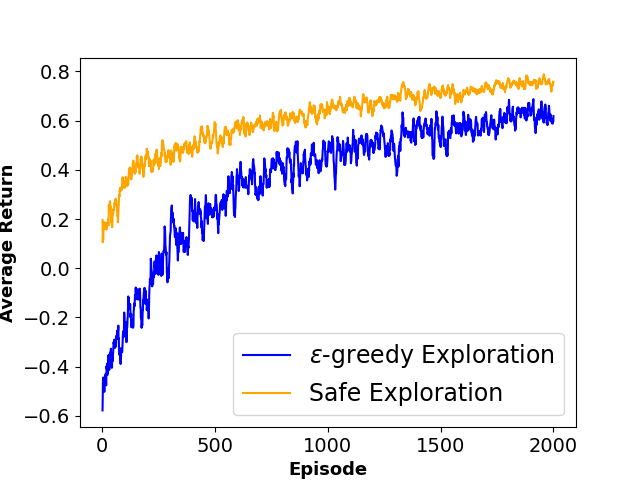}
        \caption{Task 1}
        \label{fig:sub1}
    \end{subfigure}
    \begin{subfigure}[b]{0.235\textwidth}
        \centering
        \includegraphics[width=\textwidth]{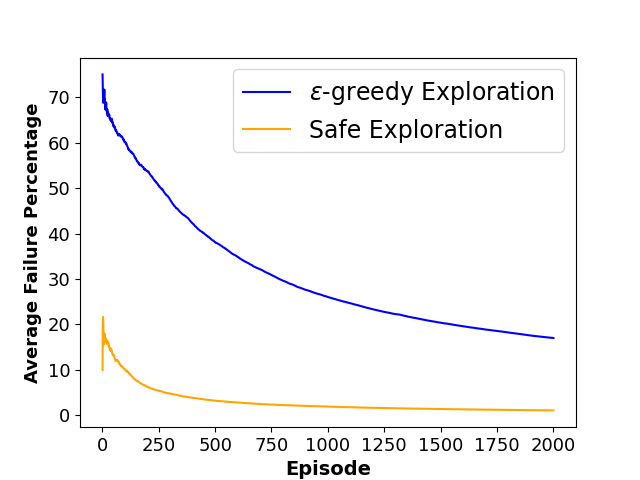}
        \caption{Task 1}
        \label{fig:sub2}
    \end{subfigure}
    \\
    \begin{subfigure}[b]{0.235\textwidth}
        \centering
        \includegraphics[width=\textwidth]{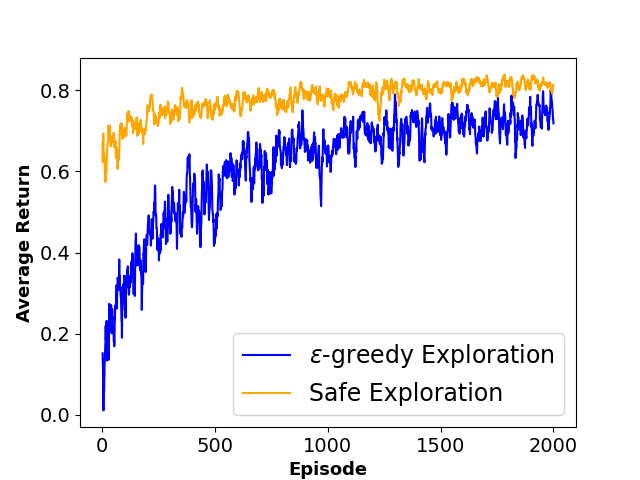}
        \caption{Task 2}
        \label{fig:sub3}
    \end{subfigure}
    \begin{subfigure}[b]{0.235\textwidth}
        \centering
        \includegraphics[width=\textwidth]{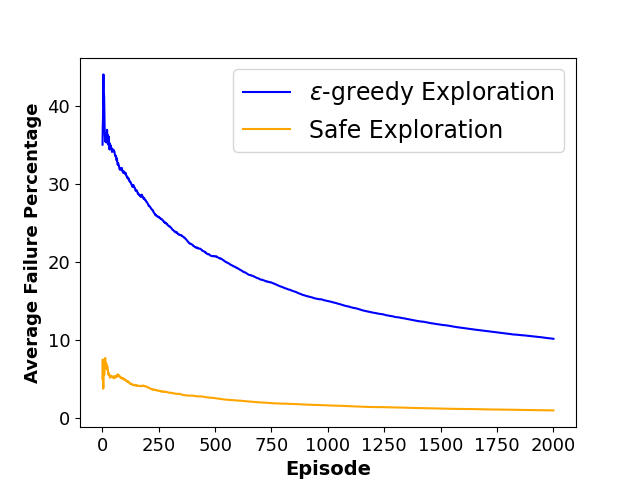}
        \caption{Task 2}
        \label{fig:sub4}
    \end{subfigure}
    \\
    \begin{subfigure}[b]{0.235\textwidth}
        \centering
        \includegraphics[width=\textwidth]{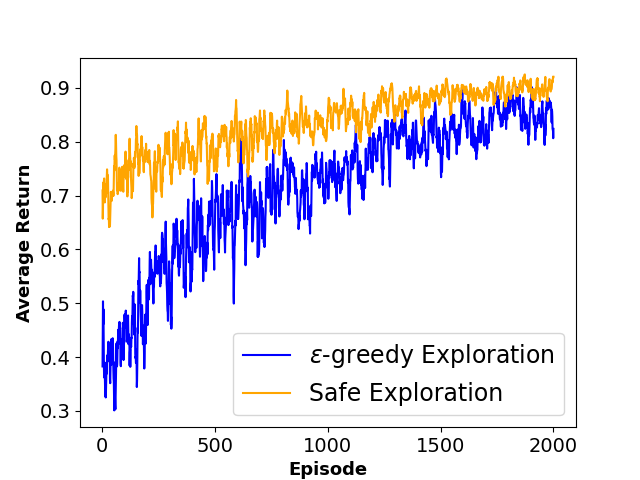}
        \caption{Task 3}
        \label{fig:sub5}
    \end{subfigure}
    \begin{subfigure}[b]{0.235\textwidth}
        \centering
        \includegraphics[width=\textwidth]{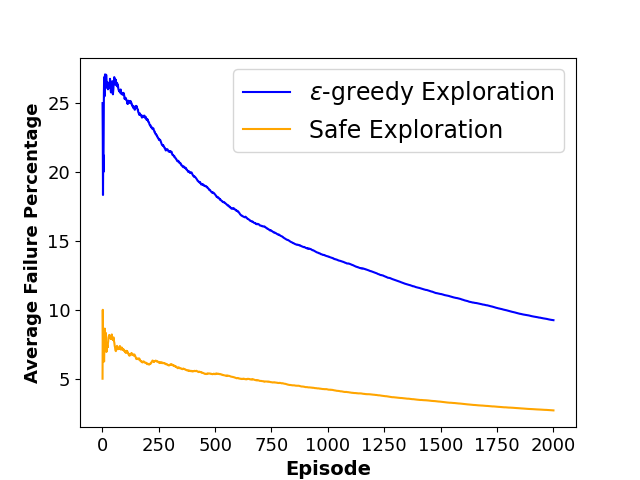}
        \caption{Task 3}
        \label{fig:sub6}
    \end{subfigure}
    \caption{Training process of SARSA algorithm for the designed tasks. Similar to Figure 3, diagrams on the left depict the mean of the average returns for the 10 most recent episodes, and those on the right illustrate the percentage of episodes that ended with collision. All values are obtained by running each test 20 times and getting the average results. We used similar hyperparameters that were used for QLearning algorithm.}
    \label{fig:overal2}
\end{figure}

\subsection{Task Adoption Using BRS Detection Model and a Safe Policy}
In the next step, we randomly generate three 15x15 grid environments, each containing a goal state, 5 blocked states that are inaccessible to the agent, and a vertically moving obstacle. To ensure the agent navigates past the moving obstacle, we randomly place the goal state in the right one-third of the environment, while placing the obstacle randomly outside that area.

We call each environment a task and train model-free agents using our framework and  $\epsilon$-greedy exploration strategy as a baseline with the same exploration rate (0.2) for 2000 episodes and 20 runs for each task. We then evaluate how safely each agent adapts to a new task in an environment that has been slightly changed.

In the proposed framework, when the agent interacts with the environment, it uses the BRS detection model to determine if the current state is safe for random exploration or using a suboptimal policy. When a state is identified as BRS, the agent follows a predefined policy focused on collision prevention rather than reaching the goal state or finding the optimal policy. Our safe policy keeps the agent in its current state if the moving obstacle is approaching by choosing "do nothing" action. If the agent is in the path of the obstacle, it considers the shortest trajectory of actions to move away from the obstacle's path, then maintains the agent's position outside the obstacle's path by selecting "do nothing" action as long as the state remains identified as BRS by the BRS detection model.

To evaluate the performance of each approach, we define three metrics namely "Average Collision Rate", representing the average percentage of episodes that concluded in a collision with a moving obstacle, "Average Success Rate", indicating the average percentage of episodes where the agent successfully reached the goal state within the defined limited time steps and "Sum of Reward", showing the sum of discounted rewards achieved by the agent throughout the training phase. The average results of agent behavior during the training process for both algorithms along with the utilized hyperparameters are also presented in Figures \ref{fig:overal1} and \ref{fig:overal2}.
 
As shown in Table \ref{tab1}, results indicate that our framework significantly reduces the number of collisions and achieves a higher reward during the training phase, which shows the effectiveness of our method in creating a safer and more efficient training process for model-free RL agents.

\section{Discussion and Future Works}
\label{sec: Discussion}
We propose a framework to enhance the efficiency of model-free agents exploring new safety-constrained grid environments. In this section, we highlight a few limitations and considerations to be aware of when utilizing this framework for other domains.

Our framework offers insights into potentially unsafe states for the model-free agent. The primary contribution of our work lies in detecting instances where random exploration or suboptimal policies may pose risks in safety-constrained environments and should be substituted with more reliable approaches. However, defining a suitable safe policy depends on the specific environment and the task. In this work, to facilitate defining an appropriate safe policy, we narrowed the scope of the work to grid environments. While we demonstrate an instance of a safe policy as a set of predefined behaviors in our experiments, defining a reliable safe behavior can be challenging for certain environments, potentially diminishing the effectiveness of this framework.

One factor that may affect the efficacy of our method is the feasibility of designing an appropriate pre-training environment for the agent. In our problem of interest, navigating the grid environments, our pre-training environment comprised the similar dynamics of the main environments. However, in many real-world scenarios, designing an ideal pre-training environment may not be feasible due to the complexity and uncertainty of these environments. Therefore, it is crucial to consider this factor before deploying the proposed framework for a task.

Another limitation of our framework arises from the nature of training a machine learning model to detect unsafe situations and the inherent risk of false detections. Therefore, this framework does not guarantee error-free recognition. Consequently, in highly sensitive domains such as medical applications, relying solely on this framework may not be advisable.

Taking into account the mentioned limitations, our framework is most effective for tasks where we can design a pre-training zone with properties similar to the real environment and define a reliable safe policy. As a few examples of potential real-world applications of our framework, we can mention the task of indoor and outdoor autonomous navigation using RL \citep{surmann2020deep, shabbir2018survey} and teaching new skills to humanoid robots \citep{garcia2020teaching}.

In future works, we aim to explore alternative approaches for defining reliable safe policies in order to enable model-free agents to devise a set of actions for each unique situation, rather than solely relying on predefined behaviors. Additionally, we aim to work on more complex environments and evaluate the effectiveness of our framework in environments with high dimensions and complex dynamics.

\section{Conclusion}
In this study, we introduce an exploration framework to enable model-free RL agents to explore new environments and adapt to new tasks more safely compared to the traditional $\epsilon$-greedy strategy. The core component of our framework is the pre-training phase that enables the agent to detect BRS states given high-level features extracted from the agent's observation by training a binary classification model. Then, this model is used in a new environment to detect whether employing random exploration or a suboptimal policy is safe. Our experimental results demonstrate that by defining an effective safe policy and pre-training the agent in an appropriate pre-training environment, the agent can learn the optimal policy in new environments with significantly fewer violations of safety constraints and higher cumulative discounted reward.
\bibliographystyle{abbrv}
\bibliography{refs}
\end{document}